\documentclass[journal]{IEEEtran}

\usepackage{amsmath}
\usepackage{graphicx}
\usepackage{subfig}
\usepackage{color}

\usepackage{siunitx}
\usepackage{microtype,textcomp}
\usepackage{wasysym}
\usepackage{cite}
\DeclareSIUnit\framespersecond{fps}

\ifCLASSINFOpdf
\else
\fi

\begin{document}

\title{Intraoperative robotic-assisted large-area high-speed microscopic imaging and intervention}

\author{Petros Giataganas$^{1}$, Michael Hughes$^{1,2}$, Christopher J. Payne$^{1}$, \\Piyamate Wisanuvej$^{1,3}$, Burak Temelkuran$^1$, 
        and~Guang-Zhong Yang$^1$,~\IEEEmembership{Fellow,~IEEE}
\thanks{$^1$Authors are with the Hamlyn Centre for Robotic Surgery, Imperial College London, SW7 2AZ, London, UK.
$^2$M. Hughes is now with the Applied Optics Group, University of Kent, UK.
$^3$P. Wisanuvej is also with Kasetsart University, Bangkok, Thailand.
E-mail: petgiat@gmail.com}
\thanks{
}}

\markboth{
} 
{Shell \MakeLowercase{\textit{et al.}}: Intraoperative robotic-assisted large-area high-speed microscopic imaging and intervention}

\maketitle

\begin{abstract}
\textit{Objective}: Probe-based confocal endomicroscopy is an emerging high-magnification optical imaging technique that provides \textit{in~vivo} and \textit{in~situ} cellular-level imaging for real-time assessment of tissue pathology. Endomicroscopy could potentially be used for intraoperative surgical guidance, but it is challenging to assess a surgical site using individual microscopic images due to the limited field-of-view and difficulties associated with manually manipulating the probe. \textit{Methods}: In this paper, a novel robotic device for large-area endomicroscopy imaging is proposed, demonstrating a rapid, but highly accurate, scanning mechanism with image-based motion control which is able to generate histology-like endomicroscopy mosaics. The device also includes, for the first time in robotic-assisted endomicroscopy, the capability to ablate tissue without the need for an additional tool. \textit{Results}: The device achieves pre-programmed trajectories with positioning accuracy of less than \SI{30}{\micro\metre}, while the image-based approach demonstrated that it can suppress random motion disturbances up to \SI{1.25}{\milli\metre\per\second}. Mosaics are presented from a range of \textit{ex vivo} human and animal tissues, over areas of more than \SI{3}{\milli\metre\squared}, scanned in approximate \SI{10}{\second}. \textit{Conclusion}: This work demonstrates the potential of the proposed instrument to generate large-area, high-resolution microscopic images for intraoperative tissue identification and margin assessment. \textit{Significance}: This approach presents an important alternative to current histology techniques, significantly reducing the tissue assessment time, while simultaneously providing the capability to mark and ablate suspicious areas intraoperatively. 
\end{abstract}

\begin{IEEEkeywords}
optical imaging, medical robotics, visual servoing, endomicroscopy.
\end{IEEEkeywords}

\IEEEpeerreviewmaketitle

\section{Introduction}

\IEEEPARstart{A}{dvances} in surgery have had a significant impact on cancer treatment in recent years, and the development of intraoperative image guidance techniques has the potential to further revolutionise the field. However, post-operative histopathological examination remains  the current `gold standard' approach by which completeness of tumour resection is confirmed. Identification of `positive' or `close' margins in histology - the presence of cancerous tissue on or near the surface of the excised specimen - is strongly associated with recurrence \cite{Batsakis1999}. Unfortunately, due to the long timescales associated with histological reporting \cite{Liu2011a}, this information often comes too late to guide an intervention. The cryosection technique (`frozen section') can be used to obtain intraoperative histological information, but it is expensive and its reliability can be affected by freezing artefacts \cite{Pleijhuis2009}. It also severely disrupts the surgical workflow, with processing times as long as \SIrange[range-units = single,range-phrase = --]{20}{30}{\minute}. Biopsy also inherently relies on discrete sampling of a given area, making it strongly dependent on the operator and potentially unrepresentative of the true state of the tissue.

Probe-based confocal laser endomicroscopy is an emerging optical imaging technique that has shown potential for intraoperative tissue characterisation. It provides real-time, \textit{in situ} visualisation of fluorescently-labelled human tissue with cellular-scale resolution, a procedure known as `optical biopsy'. These high resolution microscopy images can be acquired at video rates by relaying light to and from the tissue using fibre-optic imaging bundles and a micro-lens, allowing a compact probe design without complex mechanical components \cite{Gmitro1993}. Preliminary studies have shown potential applications in neurosurgery \cite{Zehri2014} for optimising resection, in head and neck surgery \cite{Pogorzelski2012} for early detection and resection of squamous cell carcinomas, in breast surgery \cite{Chang2015,DePalma2015} for distinguishing neoplastic from non-neoplastic cells, and for diagnostic purposes in gastrointestinal, colorectal, gastric, urinary tract, ovarian and lung cancer \cite{Jabbour2012}.
\begin{figure}[t!]
\centering
\includegraphics[width=\linewidth]{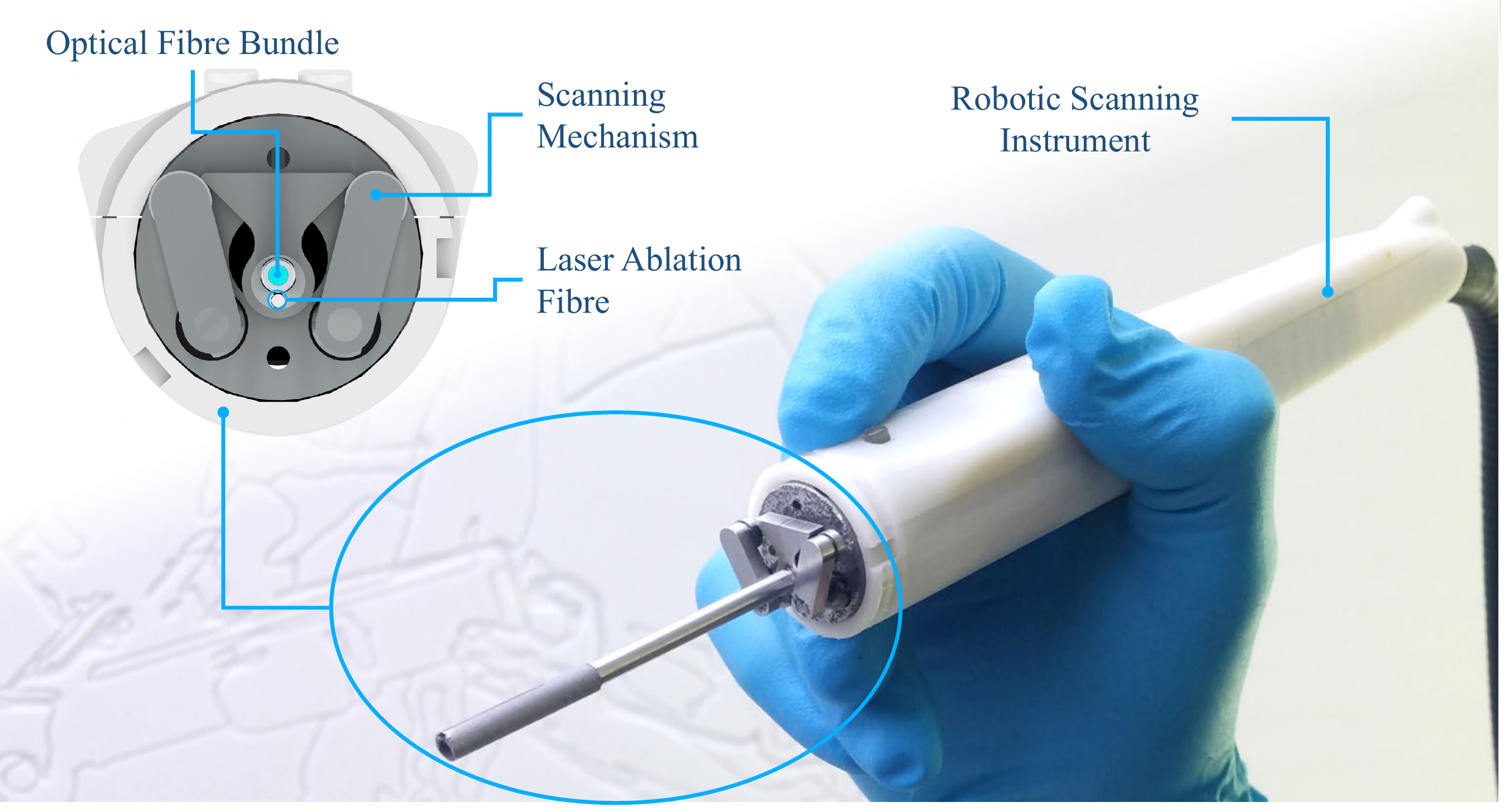}
\caption{The high-speed robotic scanning device and its individual components featuring the scanning mechanism, the incorporated fibre bundle and laser ablation fibre.}
\label{fig:fabricatedHandheld}
\end{figure}

Despite promising initial clinical trials, a significant limitation of endomicroscopy is that the field-of-view (FoV) tends to be smaller than a typical histology section, being typically only \SI{0.25}{\milli\metre} for high resolution probes. This limitation arises due to the finite number and spacing of cores in the fibre bundle, and makes it difficult for clinicians to correctly interpret tissue morphology \cite{Patsias2014}. To alleviate this issue and synthesise a broader FoV, it is necessary to stitch together (mosaic) individual image frames as the probe is moved across the tissue \cite{Vercauteren2007}. However, obtaining consistent mosaics \textit{in vivo} remains a challenge. Whilst one dimensional linear mosaics can be formed by manually manipulating the probe, it is practically difficult to scan over two dimensional areas due to the likelihood of inconsistent velocities and poorly controlled trajectories. 

Robotic manipulators and hand-held smart devices \cite{Payne2014,Yang2012d} have significant potential to provide precise probe positioning that will enhance the usability of these optical probes and improve their diagnostic yield. Initial research in robotic-assisted endomicroscopy by Latt et al. \cite{Latt2011a, TunLatt2012} demonstrated hand-held rigid instruments with force sensing capabilities. Since endomicroscopy requires the probe to be in contact with the tissue, the forces applied by the probe affect the image quality through tissue deformation \cite{Newton2011}, especially when under manual manipulation. While Latt et al. demonstrated the benefits of constant contact forces on the quality of individual images, they did not extend this to large area scanning. 

\begin{figure}[t!]
\centering
\includegraphics[width=1\linewidth]{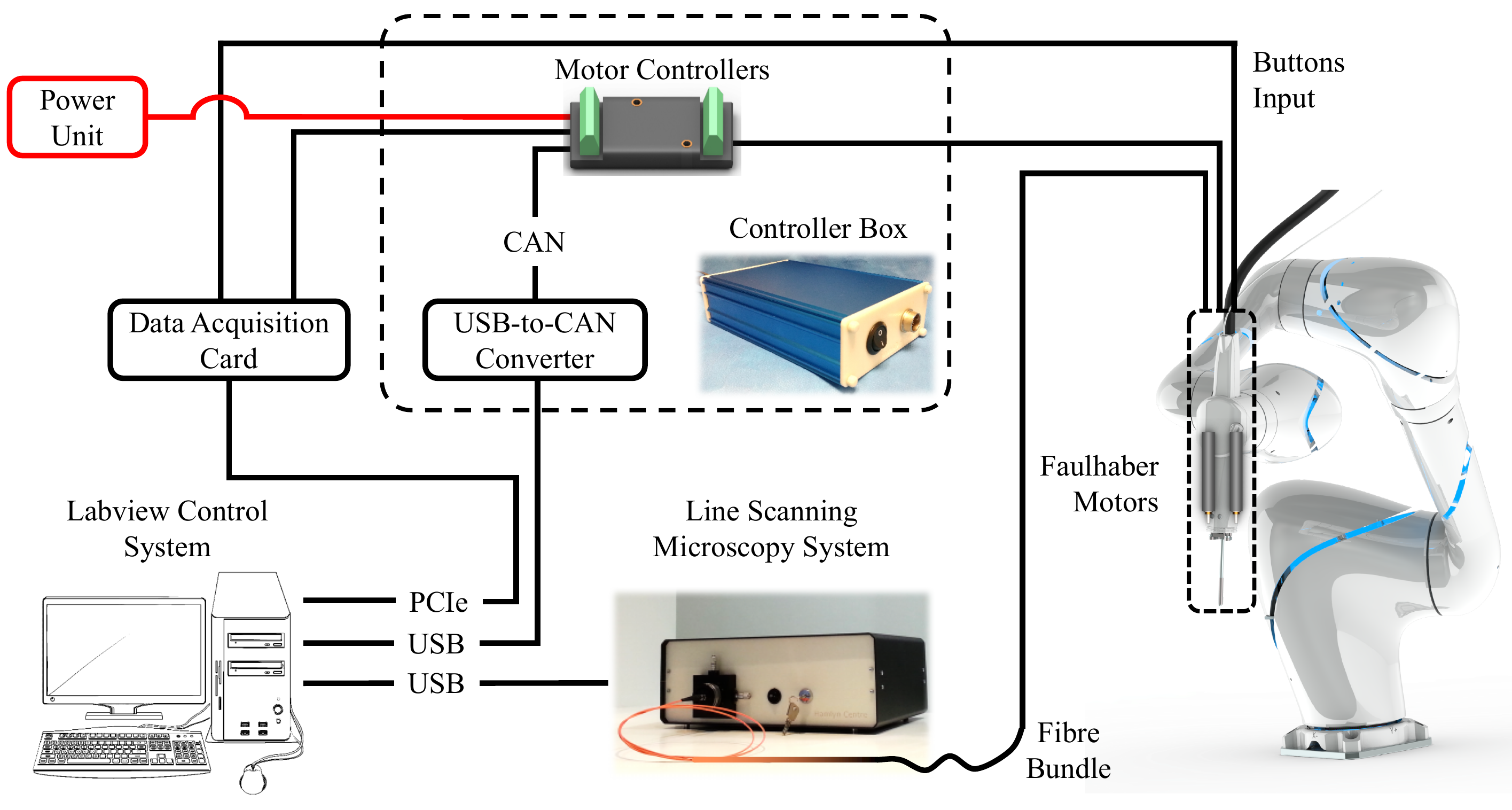}
\caption{Framework overview of the proposed system.}
\label{fig:Framework-overview-of}
\end{figure}

Combined force control and mosaicking was demonstrated  in subsequent work \cite{Giataganas2013} where a robotic manipulator was used to automatically perform large area coverage with spiral mosaics. In parallel, Rosa et al. \cite{Rosa2012} used a similar, large industrial robotic arm to explore the idea of visual servoing of the robot, using the endomicroscope images to compensate for tissue deformation during scanning. Recently, Zhang \textit{et al.} \cite{Zhang2017} used the Da Vinci surgical system to autonomously scan an endomicroscope over a predefined local area using both macroscopic and microscopic visual servoing. These approaches demonstrated the potential of robotic-assisted endomicroscopy, but employed large, complex and expensive robotic systems which are unsuitable for the majority of clinical procedures. Some of these devices also have intrinsically low accuracy (\textgreater \SI{0.2}{\milli\metre}) because they function over large workspaces, and in some cases will have relatively low bandwidth.

A series of bespoke robot-assisted endomicroscopy devices have since been developed with the aim of generating large area mosaics through minimally-invasive incisions. Newton et al. \cite{Newton2012a} showed that a 7-degrees-of-freedom (DoF) articulated robotic device offered sufficient dexterity and stability to create \SI{2}{\milli\metre} long linear mosaics with consistent image quality during peritoneoscopy in a porcine model. Giataganas et al. \cite{Giataganas2014} developed an articulated robotic device for use in transanal endoscopic microsurgery, while Rosa et al. \cite{Rosa2011, Erden2014b} demonstrated a conic mechanical scanner which could generate spiral trajectories for laparoscopic surgery, covering an area of around \SI{3}{\milli\metre\squared}. The latter, however, suffers from actuation problems in the central region and stabilisation of the tissue must be performed using an outer tube (i.e. a mechanical stabiliser). The speed of the scan was also limited by the frame rate of the endomicroscope (Cellvizio , Mauna Kea Technologies) which could generate confocal fluorescence images with a \SI{0.24}{\milli\metre} FoV only at a rate of \SI{12}{\framespersecond}.

A different approach was pursued by Zuo et al. who developed three generations of scanning devices \cite{Zuo2015, Zuo2015a} designed to provide very large area scanning within the cavity created by wide, local excision of a tumour in breast conserving surgery. These devices relied on a balloon which was to be inflated within the cavity to allow a 2-DoF probe to scan a spiral or raster pattern within a hemispherical region. While the system demonstrated significant advances in the field of large area scanning, achieving coverage of around \SI{35}{\milli\metre\squared} in \textit{ex vivo} experiments, there are a number of issues affecting the clinical uptake of this device, mainly due to open loop control of the probe and difficulty of catering for highly irregular surfaces. 

In this paper, a new robotic endomicroscopy scanning system, shown in Fig.~\ref{fig:fabricatedHandheld}, is proposed and validated for use in minimally invasive procedures. Rather than attempting whole-cavity scanning, the focus of this work is to provide the surgeon with a tool for rapid assessment of tissue micro-structure that can be integrated into the normal surgical workflow. By adopting a novel and highly repeatable method of scanning, a rigid probe with micrometer-scale accuracy over a mosaicking workspace of \SI{10}{\milli\metre\squared} is possible, representing a significant improvement on prior work \cite{Zhang2017, Newton2012a, Giataganas2014, Rosa2011, Erden2014b, Zuo2015, Zuo2015a}. The system is demonstrated together with a high frame rate endomicroscopy system that significantly reduces scanning time compared with the current state-of-the-art. A visual servoing algorithm, making use of the endomicroscope images for real-time scanning trajectory correction, allows generation of mosaics even over highly-deforming tissue structures. The system is demonstrated on \textit{ex vivo} human breast tissues, with a view for it to be used in breast conserving surgery, although the platform is generally applicable to a range of surgical procedures. Additionally, for the first time in robotic-assisted endomicroscopy, we provide an initial demonstration that energy delivery (a CO$_{2}$ laser in our example) can be incorporated into the scanning device, either for image-guided ablation or marking tumour margins. 

\section{Materials and Methods}

\subsubsection{Robotic Scanning System}

The intraoperative scanning endomicroscopy system, shown in Fig.~\ref{fig:fabricatedHandheld}, consists of the endomicroscope, a scanning instrument, a fibred ablation laser and (optionally) an auxiliary robotic arm. An overview of the system components is provided in Fig.~\ref{fig:Framework-overview-of}. The instrument's scanning mechanism allows the endomicroscopy probe to be moved over an area of several mm while individual image frames are stitched together to form a mosaic. It is enclosed in an ergonomic casing for hand-held operation, with two buttons to control the scan during hand-held operation. A \SI{58}{\milli\metre} long steel tube of \diameter{\SI{3.3}{\milli\metre}} outer diameter and \diameter{\SI{2.7}{\milli\metre}} inner diameter acts as a channel for passing multiple fibres (Fig.~\ref{fig:fabricatedHandheld}(a)). 

\begin{figure}[t!]
\centering
\includegraphics[width=1\linewidth]{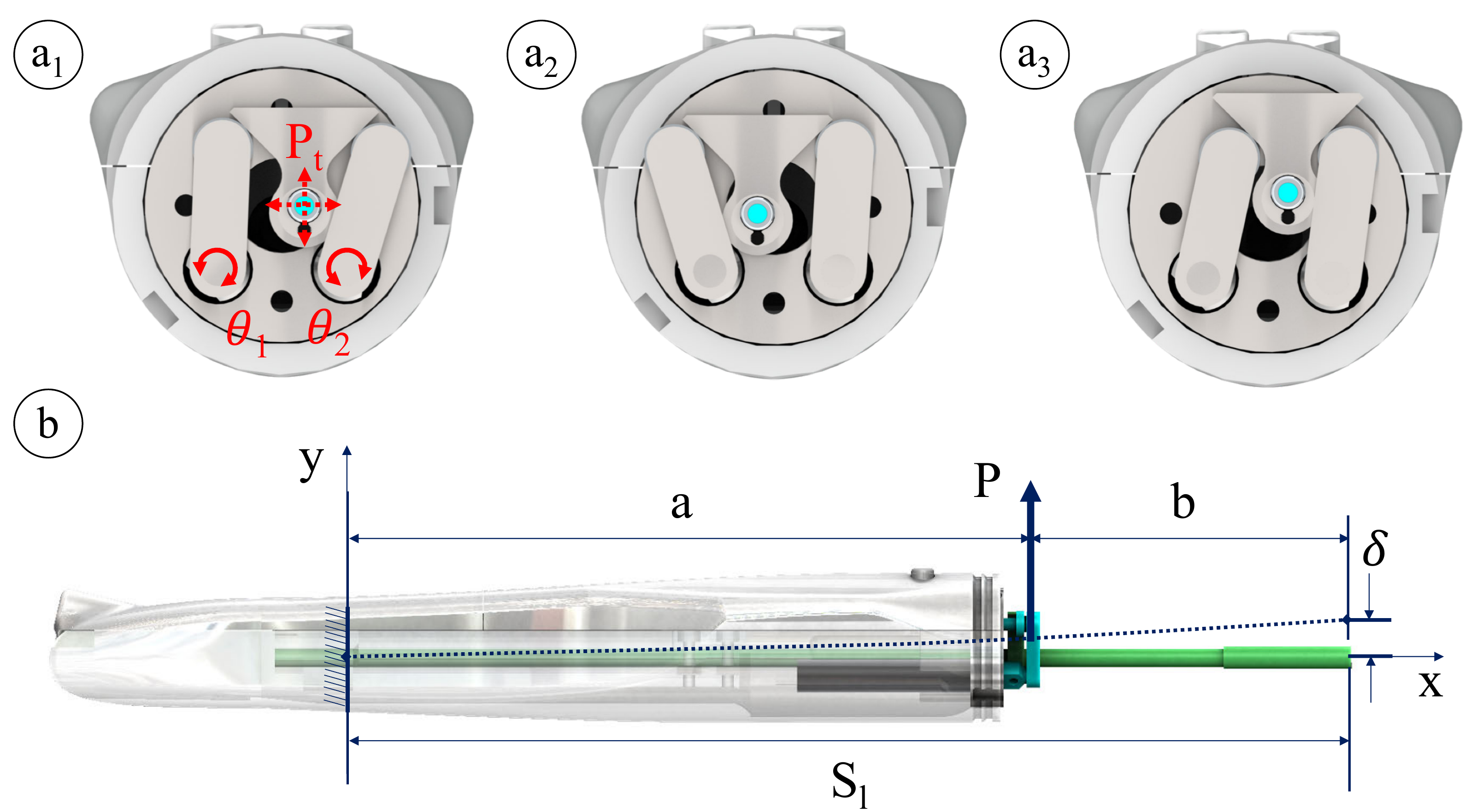}
\caption{ (a) Three scanning positions presenting three different steel level positions with the resulting deflection on the device's shaft. The two motor angles $\theta_1$, $\theta_2$ and the resulting tip position $P_t$ are presented. (b) Side view of the instrument presenting the shaft (green), the tube deflection in a random position under load P (blue) and the shaft's fixation point.}
\label{fig:mechanicalOverview}
\end{figure}

The steel tube is clamped in a cantilever configuration so that it can deflect in two planes at its free end. Two cam-roller mechanisms are then used in conjunction with micro servomotors to deflect the tube, as shown in Fig.~\ref{fig:mechanicalOverview}.  A V-profiled steel cam is welded to the tube shaft; it is engaged by two steel levers with tip mounted bearings that exert lateral forces on the cantilevered tube, with the levers actuated by two \diameter{\SI{6}{\milli\metre}} micro servomotors (Brushless DC-Servomotor 0620, Faulhaber, Germany) with a 256:1 reduction gearbox (Faulhaber, Germany). The cantilevered tube is mounted in an unloaded position outside of the nominal workspace of the device; this ensures that the cantilevered tube is always deflected and thus preloaded against the cam-roller mechanisms so as to avoid backlash. In contrast to other hand-held micro-manipulating devices that use piezo-bimorph elements or piezo-squiggle motors, the cantilever tube mechanism used here provides precise, controlled motion of the instrument tip with minimal backlash. This leads to more accurate and faster generation of larger mosaics than previous methods in a compact hand-held configuration.

The endomicroscopy probe is inserted through the steel tube which can also accommodate a second fibre for energy delivery, with both fibres fixed proximally via a locking mechanism. Other modalities of optical imaging probes could also be passed through this channel, providing the surgeon with a scanning platform for various optical biopsy techniques. An exploded view of the instrument is provided in Supplementary Fig.~S1.

The motor controllers' embedded software communicates with the control system in two ways. A CAN bus (Faulhaber CANopen interface) with 1000\:kbps Baudrate is used for the initialisation process and powering on/off the motors. The main kinematic control, however, is performed through analog voltage signals using the analog input of the motor controllers, providing a faster update rate than the CAN bus (\SI{200}{\hertz}). The analog voltage input is provided through a data acquisition card (PCIe-6321, National Instruments, USA) that communicates through a Labview interface and a PCI Express card (\SI{1}{\kilo\hertz}). All components are powered with a \SI{6}{\volt} power supply, which is controlled by a switch integrated on the controller box. 

Closed-loop position control of the servomotors uses an integrated magnetic encoding system (analogue hall sensors (K2280, Faulhaber, Germany)) and two dedicated motor controllers (MCBL 3002 S CF, Faulhaber, Germany), which also provide power to the motors. The cantilevered tube and micro servomotors are mounted in a tubular chassis, enclosed by the outer casing. The tubular chassis was fabricated with Veroblack material using a additive manufacturing machine (Objet EDEN350$^{TM}$, Stratasys Ltd., USA) while the casing was fabricated with ABS M30i thermoplastic material (Fortus 400mc, Stratasys Ltd., USA). The motor controllers are placed distally in a separate enclosed box to reduce the size and weight of the instrument. 

The scanning mechanism allows the probe to be driven over a usable workspace of up to approximately \SI{14}{\milli\metre\squared}. For high resolution probes, with a FOV of \SI{0.058}{\milli\metre\squared}, this represents a potential 240-fold increase in the imaging area. A spiral scan pattern was found in practice to provide the highest quality mosaics, with fewest discontinuities, but in principle any scan pattern could be employed. A CAD simulation (see Supplementary Fig.~S3,S4) shows how the configuration of the cam-roller mechanism was optimised to achieve the largest possible workspace for the smallest possible overall dimensions of the actuation system. It also shows that the axial deviation of the probe tip from a plane is less than \SI{50}{\micro\metre} at the extremes of the workspace, making loss of tissue contact unlikely when tissue elasticity is considered. 

The scanning instrument can be used hand-held to acquire individual image frames or deliver energy for tissue ablation, or alternatively used in conjunction with an external stabilising arm. The mosaics presented here were acquired with the instrument fixed in place using a stabilising arm. The operator could then manipulate the arm in a hands-on fashion to position the device to image the desired area of tissue before initiating a scan. The instrument was also tested mounted on an auxiliary robotic arm, developed by Wisanuvej et al. \cite{Wisanuvej2016,Wisanuvej2017}, as shown in Fig.~\ref{fig:fabricatedHandheld}(c). The robotic arm is a 6-DoF serial articulated robot with \SI{1.5}{\kilo\gram} maximum payload. It has integrated brushless DC motors, coupled with harmonic drive gears that, in case of power loss, allow each joint to be back-driven by the operator. In normal operation, the surgeon can manipulate the arm in a hands-on fashion by pressing a foot pedal, which when released rigidly locks the entire robotic arm.

\subsubsection{Kinematic Model Analysis}

For the kinematic analysis of the device, a two-step approach was used. Initially, using a geometric method described in the Supplementary Materials (Section 2), the two motor angles $\theta_1$ and $\theta_2$ are used to find the position $P(x,y)$ of the shaft centre with respect to the coordinate system at the origin point $O$. In the second step of the analysis, the tube deflection is used to find the end-effector position at the tip of the shaft. Initially, the Cartesian coordinates $P(x,y)$ are converted to polar coordinates $P(r_p,\theta_p)$ and we compute the end deflection based on knowing the deflection part way along the tube $\delta$, using beam theory. We then convert back to Cartesian coordinates to obtain the $P_t(x_t,y_t)$ positions at the tip. In this approach, we assume that there are no large deflections, the plane sections remain plane and the material has linear elastic behaviour. We also assume minimal z motion both at the cam deflection point and at the end deflection point. These assumptions are well within the manufacturing/assembly tolerances and hence are reasonable.

As shown in Fig.~\ref{fig:mechanicalOverview}(b), using the beam deflection formula for concentrated load $P$ at any point, the maximum deflection $\delta$ is calculated by: 
\begin{equation} \delta = \frac{Pa^2}{6EI}(3{S_l}-a)  \end{equation}
where ${S_l}$ is the length of the shaft, $E = 209$ GPa is the elastic modulus of steel, and $I$ is the second moment of area: 
\begin{equation} I =  \frac{\pi}{64}(OD^4-ID^4) \end{equation}
where $OD = 3.3$ mm is the outer diameter of the tube, and $ID = 2.7$ mm is the inner diameter, and 
\begin{equation} P =  \frac{6EIr_p}{2a^3} \end{equation}

Finally, the tip position $P_t(x_t,y_t)$ in Cartesian coordinates is given by: 

\begin{equation}  x_t = \delta sin(\theta_p)  \end{equation}
\begin{equation}  y_t = \delta cos(\theta_p)  \end{equation}

By driving the instrument in a straight line within the linear workspace, the scaling factor between analog voltage inputs and the Cartesian correspondences can be calculated using the mosaic image generated. This was found to be $664$ \si{\micro \metre \per \volt} for both directions. The correspondence between the motor rotations (in degrees) and the analog input voltages can be found from the following motor settings: (a) voltage range $= 20\,V$, (b) motor increments range (without gear transmission) $= 100000$, (c) motor increments range (without gear transmission) per revolution $= 3000$ and (d) gear ratio $= 256$. Hence, the value of R (in degrees per volt) can be determined to be:

\begin{equation} R = 360 \frac{1}{256} \frac{1}{3000} \frac{100000}{20} = 2.3438 ^o/V
\end{equation}

\subsubsection{Optical Imaging Systems}

The primary imaging system used with the robotic instrument is an in-house fluorescence virtual slit-scanning endomicroscopy system that allows high frame rate imaging and hence higher-speed robotic scanning. An optical fibre bundle based probe (Gastroflex UHD, Cellvizio, Mauna Kea Technologies) is used, which incorporates a Fujikura imaging bundle (core spacing  \SI{3}{\micro\metre} and \SI{30000}{cores}) and an approximately x2.5 magnification distal micro-objective, providing a \SI{240}{\micro\metre} FOV and a resolution of approximately \SI{2}{\micro\metre}. The optical system is fully described in Hughes et al. \cite{Hughes2016} and only a brief overview is presented here. A galvanometer mirror is used to scan a line-shaped \SI{488}{\nano\metre} laser beam over the proximal end of the fibre bundle; the bundle transfers this to tissue, exciting a line of fluorescently stained tissue. The excited fluorescence is relayed through the bundle, wavelength-filtered to remove reflected illumination light and imaged onto a CMOS camera (Point Grey Flea 3, FL3-U3-12S2M-CS) with a rolling shutter. The rolling shutter of the camera is synchronised with the line scanning, providing optical sectioning at \SI{120}{\framespersecond} without a 2D scanning system. 

Real-time processing is provided by a Labview interface and includes a Gaussian spatial filter of \SI{1.4}{px} to reduce artefacts introduced by the `honeycomb pattern' of the fibre cores, a circular mask to remove the edges of the bundle, and subtraction of a darkfield background image to remove fluorescent signal from the fibre bundle. 

Other optical imaging systems can be used with the proposed robotic scanning device providing the probe fits through the \diameter{\SI{2.7}{\milli\metre}} diameter tube channel. The scanner has been tested with an in-house dual wavelength slit scanning system, running at \SI{60}{fps} (of a similar design as the single wavelength version) and an in-house endocytoscopy system \cite{Hughes2013}. The endocytoscope generates colour endomicroscopy images at \SI{15}{fps} with a FOV of \SI{600}{\micro\metre} and a resolution of approximately \SI{6}{\micro\metre}. Images and mosaics from all systems are presented in the results section.  

\subsubsection{Mosaicking}

The real-time mosaicking algorithm is similar to previously reported approaches \cite{Vercauteren2008a} and can run at the full-frame rate of the endomicroscopy system (\SI{120}{\framespersecond}). Normalised cross correlation (NCC) is used to determine the shift between two successive images $I_1$ and $I_2$. After images are resized to a \SI{200}{px} diameter, a central template is extracted from $I_1$ (here \SI{75}{px} in size) and the cross-correlation is found with $I_2$. The peak of the cross-correlation is taken to be the shift between the two images. A running total is kept of the current position (by integrating the shifts), and the frame is added (deal-leaf) to the mosaic image at the estimated position. The algorithm was integrated with the Labview environment in the form of a dynamic link library (DLL), written in C++ using the openCV library. 

\subsubsection{Trajectory Generation}

The device can scan arbitrary trajectories. Practical scanning trajectories require a constant velocity  (linear or tangential) to maintain a consistent overlap between consecutive endomicroscopy images. For example, to generate a linear scan, the operator specifies the scan point frequency $f_{s}$, the linear velocity $u_{s}$ and the trajectory length $l_{s}$. Based on these inputs, each point $({x_v}_{(i)},{y_v}_{(i)})$ is generated as: 
\begin{equation}
({x_v}_{(i)},{y_v}_{(i)})=({x_{v_c}}+(\frac{i\cdot u_{s}}{f_{s}}),{y_{v_c}})
\end{equation}
where $i=\{0,n_p\}$, $n_p$ is the number of points, and $({x_{v_c}},{y_{v_c}})$ is the initial centre point used as offset to the trajectory. The same principle can be used for generating raster scan trajectories.  For practical scanning, to minimise deformation due to sharp turns of the probe, we used points equally spaced along an Archimedean spiral trajectory, following a previously described approach \cite{Zuo2015}. Note that, due to the approximated spiral length, near the centre of the spiral the points are only approximately uniformly spaced.  Further information about the kinematic model of the device and the conversion between the analog voltage inputs and the commanded Cartesian inputs is presented in Supplementary Sect.~$2-4$. Conversion between voltage inputs and the commanded Cartesian inputs is practically linear and was confirmed using a custom tracking rig.





\subsection{Robot Closed-Loop Visual Servoing using Microscopic Images}

Since the endomicroscope probe is in constant contact with the tissue, tissue deformation tends to occur during scanning. Therefore, the resulting mosaic's shape does not match the pre-planned trajectory, \textit{i.e.} the probe's motion relative to the deforming tissue is not the same as the probe's absolute motion. To mitigate this effect, a model-free visual servoing approach was used, similar to that of Rosa \textit{et al.} \cite{Rosa2012}, making use of the positional information from the endomicroscopy images. As the instrument scans a planned trajectory, the estimated probe position of the endomicroscope relative to the tissue surface is derived from the integration of inter-frame shifts calculated by the mosaicking algorithm. These shifts are used as the input to a closed loop proportional\textendash integral\textendash derivative (PID) control system to adjust the scanning trajectory in real time so as to achieve the desired scan pattern.

The control algorithm uses the estimated probe position in the 2D mosaic image plane (in pixels) to control the $({x_v},{y_v})$ positions of the instrument. The transformation between the mosaic image space (in pixels) and the probe's position space requires experimental determination of the rotation angle $\varphi$ and the pixels-to-voltage scale factor $L$. The conversion from the mosaic image co-ordinates $({x_I},{y_I})$ to positions $p_{v}=({x_v},{y_v})$ is then given in terms of $L$ and a rotation matrix $\mathbf{R}$: 
\begin{equation}
{\scriptstyle  \left[\begin{array}{c}
x_{v}\\
y_{v} 
\end{array}\right]=L\cdot\left[\begin{array}{cc}
cos\varphi & sin\varphi\\
-sin\varphi & cos\varphi
\end{array}\right]\cdot\left[\begin{array}{c}
x_{I}\\
y_{I}
\end{array}\right]=L\cdot R\cdot\left[\begin{array}{c}
x_{I}\\
y_{I}
\end{array}\right]}
\end{equation}
$L$ is a fixed value for each probe, whereas the angle $\varphi$ depends on the probe's rotation when loaded into the scanner, and so is calculated at the beginning of each experiment by repeatedly performing linear scans in the `x' direction and adjusting $\varphi$ until the linear scan is best aligned with the mosaic `x' axis. 

At the beginning of each scan, the probe's position in the mosaic image is initialised to $p_{I}(t=0)=(0,0)$, corresponding to the centre $p_{Ic}=(x_{Ic},y_{Ic})$ of the mosaic image. During scanning, the estimated probe position $p_{v}(t)$ at  time instance $t$ is compared with the desired trajectory position $p_{v}^{*}(t')$ at time $t'$. Due to the different sampling rates between the planned trajectory and the endomicroscopy image acquisition, the closest planned time point $t'$ to the actual measurement time $t$ is used. The  PID controller then minimises the difference between $p_{v}(t)$ and $p_{v}^{*}(t')$. In practice, a PI controller is sufficient as the scanning task is quasi-static as the variables change slowly due to the approximately constant velocity \cite{Erden2013a}. The required correction to the probe trajectory is: 
\begin{equation}
\begin{split}
&\left[\begin{array}{c}
\Delta {p_v}_x\\
\Delta {p_v}_y
\end{array}\right]=K_{p}\cdot\left[\begin{array}{c}
{p_v}_x^{*}(t')-{p_v}_x(t)\\
{p_v}_y^{*}(t')-{p_v}_y(t)
\end{array}\right]+
\\
&+K_{I}\cdot\int_{0}^{t}\left(\left[\begin{array}{c}
{p_v}_x^{*}(\tau')-{p_v}_x(\tau)\\
{p_v}_y^{*}(\tau')-{p_v}_y(\tau)
\end{array}\right]\right)d\tau
\end{split}
\end{equation}
where the gains $K_{p}$ and $K_{I}$ are the proportional and integral
gains respectively. The gains were manually tuned to $K_{p}=$ \SI{10}{\per\minute}
and $K_{I}=$ \SI{0.4}{\per\square\minute} via the standard method of eliminating the steady state error and
minimising the overshoots. 

\section{Results}

\subsection{Mechanical and Workspace Evaluation of the Scanning Instrument}

\begin{figure}[t]
\centering
\includegraphics[width=1\linewidth]{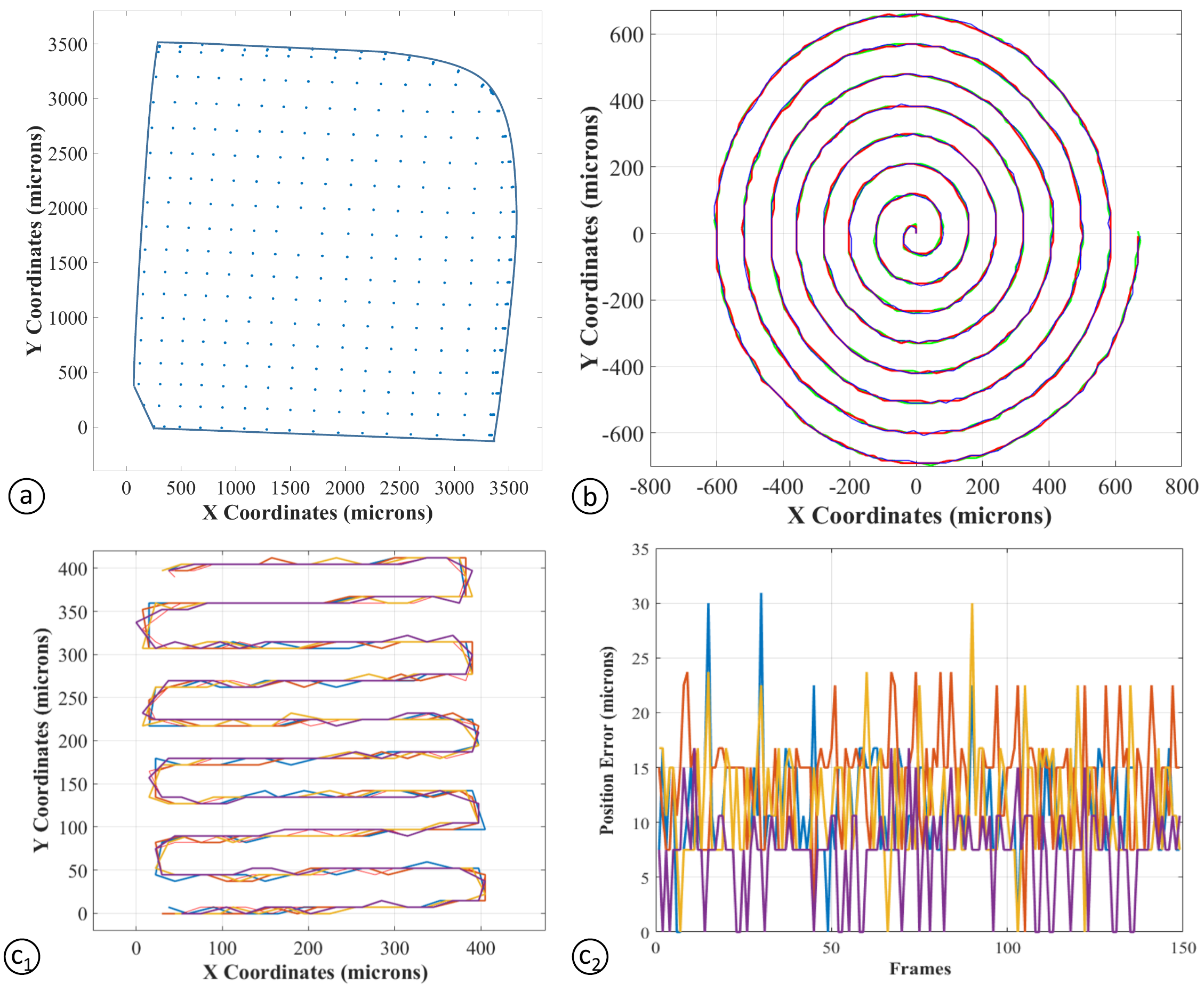}
\caption{(a) Workspace visualisation and (b-c) repeatability evaluation of two different scanning patterns. (b) Three repeats of spiral trajectories and ($c_1$) four repeats of raster trajectories. ($c_2$) Demonstration of the raster trajectory errors relative to the raster commanded trajectory.\vspace{-0.3cm}}
\label{fig:mechanical_performance}
\end{figure}

The actuation mechanism forgoes the use of any tendon or gear mechanisms, and is always under load, and thus provides high accuracy and repeatability. To validate the performance, a custom optical tracking system (Supplementary Fig.~S5) was used to track the tip of the endomicroscope during scanning. To confirm the usable workspace, a scan using incremental linearly distributed commanded positions was performed across the whole workspace, which consists of regions with linear and non-linear responses to the motor positions (see Fig.~\ref{fig:mechanical_performance}(a)). In the upper and right edges of the workspace the mechanism's levers lose contact with the cam, whereas in the lower and left edges there is a mechanical stop in the cam mechanism. The resulting central optimal linear workspace, after transforming the analog voltage values into Cartesian units, is approximately \SI{14}{\milli\metre\squared} (\num{3.7 x 3.7} \SI{}{\milli\metre} area size). Subsequently, the motion of the instrument was constrained so as to stay within this linear area. Additionally the linear distribution of these commanded positions was assessed by calculating the neighbouring points' distances. The mean distance was \SI{214}{\micro\metre} and the interquartile range was \SI{23}{\micro\metre} throughout all the workspace. These errors include geometrical errors due to the curvature of the scan plane, which are thus shown to be minimal.

The device's repeatability and accuracy were evaluated by performing two scanning patterns (spiral and raster) within the linear workspace. Fig.~\ref{fig:mechanical_performance}(b-c) and Supplementary Fig.~S6 shows the tracked trajectories over a central area with parameters typical of scans used for imaging, for three repeats of spiral trajectories and four repeats of raster trajectories. Positioning errors do not exceed \SI{30}{\micro\metre}, with a median error across all the trajectories of \SI{10.7}{\micro\metre} and an interquartile range of \SI{7.5}{\micro\metre}.

\subsection{Microscopic Visual Servoing}

\begin{figure}[t]
\centering
\includegraphics[width=0.9\linewidth]{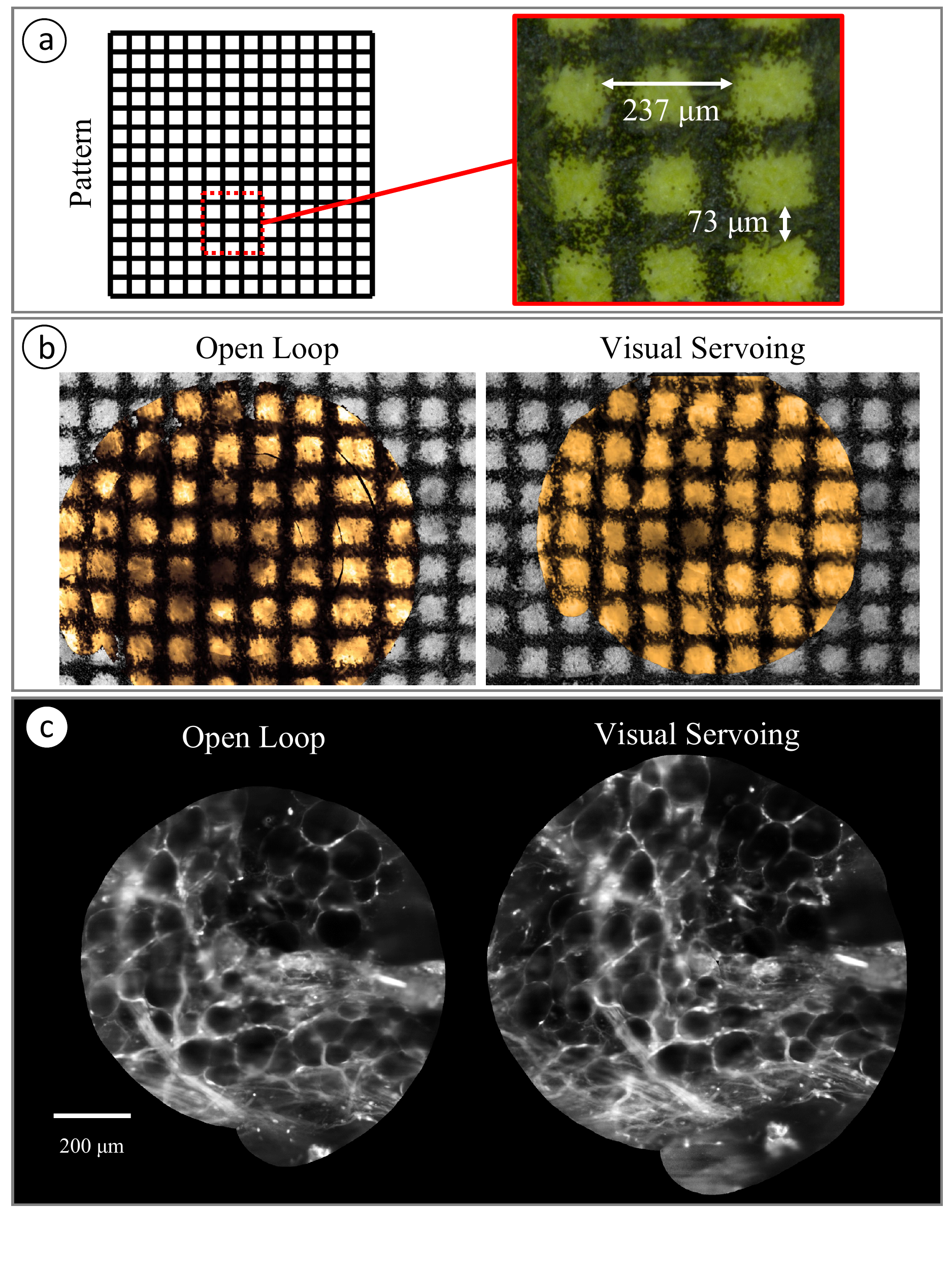}
\caption{Visual servoing results. (a) Grid pattern along with the printed pattern's microscope image, (b) two spiral mosaics (orange tint) overlaid manually onto the printed pattern's microscope image (grey) for visual comparison, and (c) mosaicking results of normal breast tissue without and with visual servoing, showing the improvement in the scanning area coverage.}
\label{fig:VisualServoingResults1}
\end{figure}

A custom phantom with a pre-defined grid pattern (see Fig.~\ref{fig:VisualServoingResults1}(a)) was used for the validation and evaluation of the microscopic visual servoing. The phantom was printed on a sheet of paper by a laser printer and coated using a fluorescent marker pen, making it visible in the fluorescence endomicroscopy images. The squares in the grid pattern had a nominal line thickness of \SI{73}{\micro\metre} and a width of \SI{237}{\micro\metre}. Using a bench-top microscope, the accuracy of the printed phantom was confirmed to be better than \SI{+-6}{\micro\metre} and \SI{+-4}{\micro\metre}. 

The phantom was scanned using both open-loop control and visual servoing; the resulting mosaics are shown in Fig.~\ref{fig:VisualServoingResults1}(b). For each mosaic image, 22 manual measurements were made of the line thickness and square widths; open-loop control resulted in a line thickness of \SI[separate-uncertainty=true]{70\pm3}{\micro\metre} and square width of \SI[separate-uncertainty=true]{236\pm13}{\micro\metre} whereas visual servoing resulted in \SI[separate-uncertainty=true]{69\pm3}{\micro\metre} and \SI[separate-uncertainty=true]{235\pm10}{\micro\metre} respectively. Both open loop and visual servoing modes reproduce the correct pattern dimensions. However, even for this non-deforming sample, and despite the high accuracy of the scanning instrument, open loop operation resulted in visually apparent errors in the reconstruction. This is partly due to the orientation of the instrument; it was not exactly perpendicular to the scanning surface, leading to a minor drift in one direction. In comparison, where visual servoing was used, there are significantly reduced visual errors.

{\color{blue}

\begin{figure}
\begin{centering}
\includegraphics[width=\columnwidth]{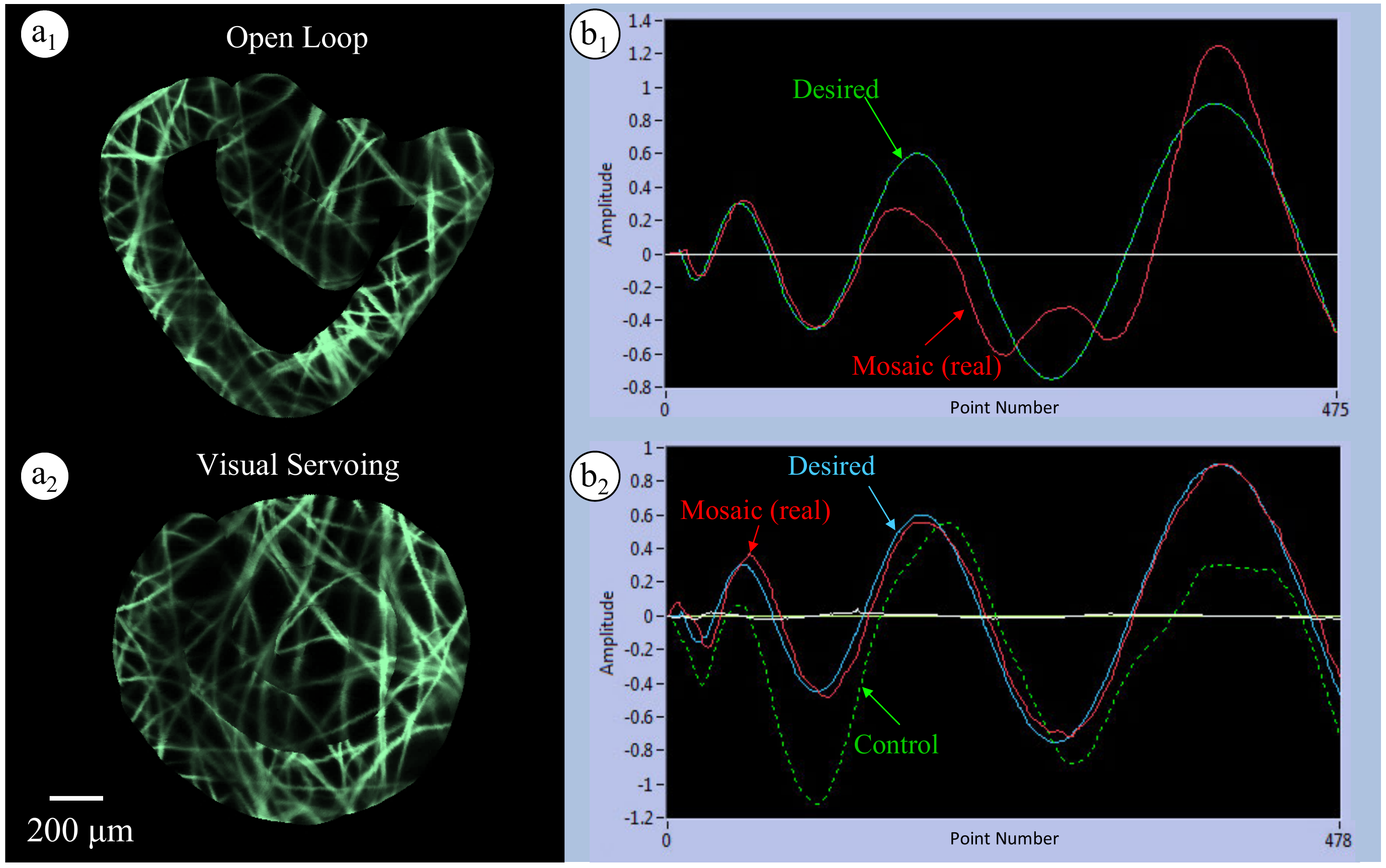}
\par\end{centering}
\caption{(a) Visual servoing mosaic results when motion is introduced using the motorised translation stage and (b) graphs showing the current position (real position from the mosaic) and the desired position for one axis only. In open loop operation, the real position deviates from the desired trajectory, whereas with the visual servoing the real position follows the desired trajectory. The control trajectory shows the analog drive signal sent to the instrument. Pseudo-colour is applied to the mosaics for improved visualisation.\label{fig:(a)-Visual-servoing}}
\end{figure}}

Fig.~\ref{fig:VisualServoingResults1}(c) compares scanning over \textit{ex vivo} human breast tissue with and without visual servoing. The tissue samples were acquired under an Imperial College tissue bank license (Project~R12047). Contrast agent acriflavine $0.02\%$  was applied topically for \SI{30}{\second} and gently washed with water to remove excess stain. The diameter of the mosaic created using visual servoing is \SI{1.1}{\milli\metre}, while the diameter reduces to \SI{0.94}{\milli\metre} when the scanner is driven open loop. This difference is due to the correction of tissue deformation by the visual servoing. Implementing a pre-planned, open-loop correction for tissue deformation would not be possible without \textit{a priori} knowledge of the tissue properties.

An even more significant difference in terms of open loop and closed loop control is observed when unexpected (\textit{i.e.} not pre-known) sample motions occur. In order to provide a simple demonstration of this, a sample of lens tissue paper was stained with  acriflavine and placed on top of a motorised translation stage. During the scan and mosaic acquisition, the stage was moved in a random pattern along both the x and y-axes (perpendicular to the axis of the scanning instrument) within a range of \SI{+-100}{\micro\metre} and with velocity of \SI[per-mode=symbol]{1}{\milli\metre\per\second}. As can be seen in Fig.~\ref{fig:(a)-Visual-servoing}(a), and as expected, the spiral trajectory failed when using open loop control, since no compensation was made for the motion of the sample. The plots in Fig.~\ref{fig:(a)-Visual-servoing}(b) show how the open-loop trajectory measured by the mosaicking algorithm, which is approximately the real trajectory followed by the probe relative to the tissue, deviates significantly from the planned trajectory (whereas it would ideally coincide). Conversely, with the visual servoing algorithm, the effects of the motion are almost eliminated as the mosaic trajectory follows the desired spiral trajectory. Note that the control trajectory, which represents the actual drive signals provided to the instrument, does not follow any specific model, demonstrating the importance of the model-free PID approach used. Tolerance of the device to the range of motion experienced in clinical practice remains a topic for further investigation.

\begin{figure}[t!]
\centering
\includegraphics[width=0.9\linewidth]{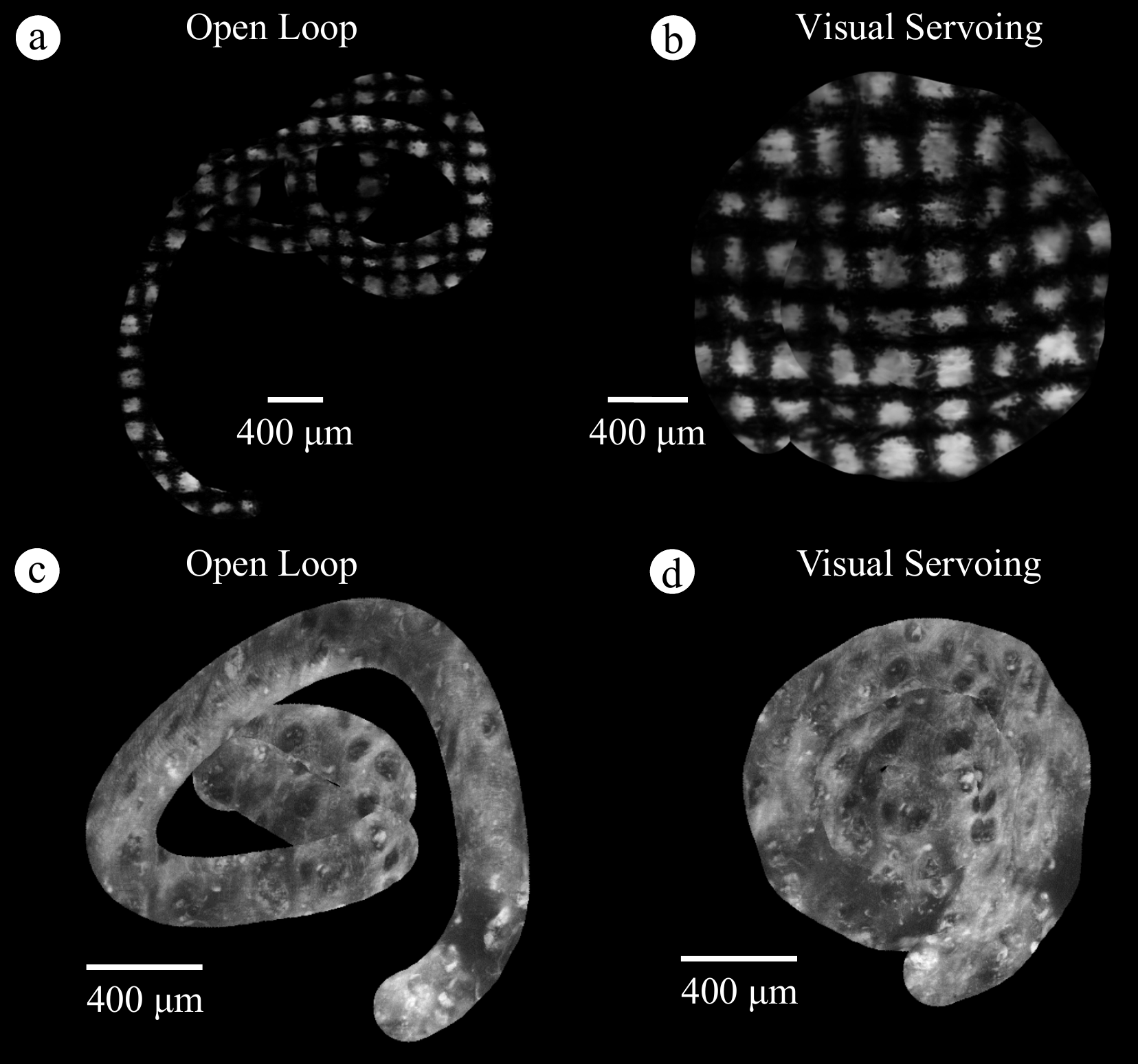}
\caption{Visually servoed mosaic results when sample motion is introduced using a motorised translation stage. (a-b) Grid pattern phantom mosaic results with random motion disturbance of \SI[per-mode=symbol]{1.25}{\milli\metre\per\second}. (c-d) Porcine colon tissue with random motion disturbance of \SI[per-mode=symbol]{0.1}{\milli\metre\per\second}.}
\label{fig:Visual-servoing-results}
\end{figure}

Further examples of the effects of external disturbance (in the form of random motion) are shown while scanning over the grid pattern in Fig.~\ref{fig:Visual-servoing-results}(a-b) and over \textit{ex vivo} porcine colonic tissue stained with acriflavine in Fig.~\ref{fig:Visual-servoing-results}(c-d). Note here that the colonic tissue results show not only the external motion's suppression but also compensation for tissue deformation during the scan. 

\begin{figure}[t!]
\centering
\includegraphics[width=0.9\linewidth]{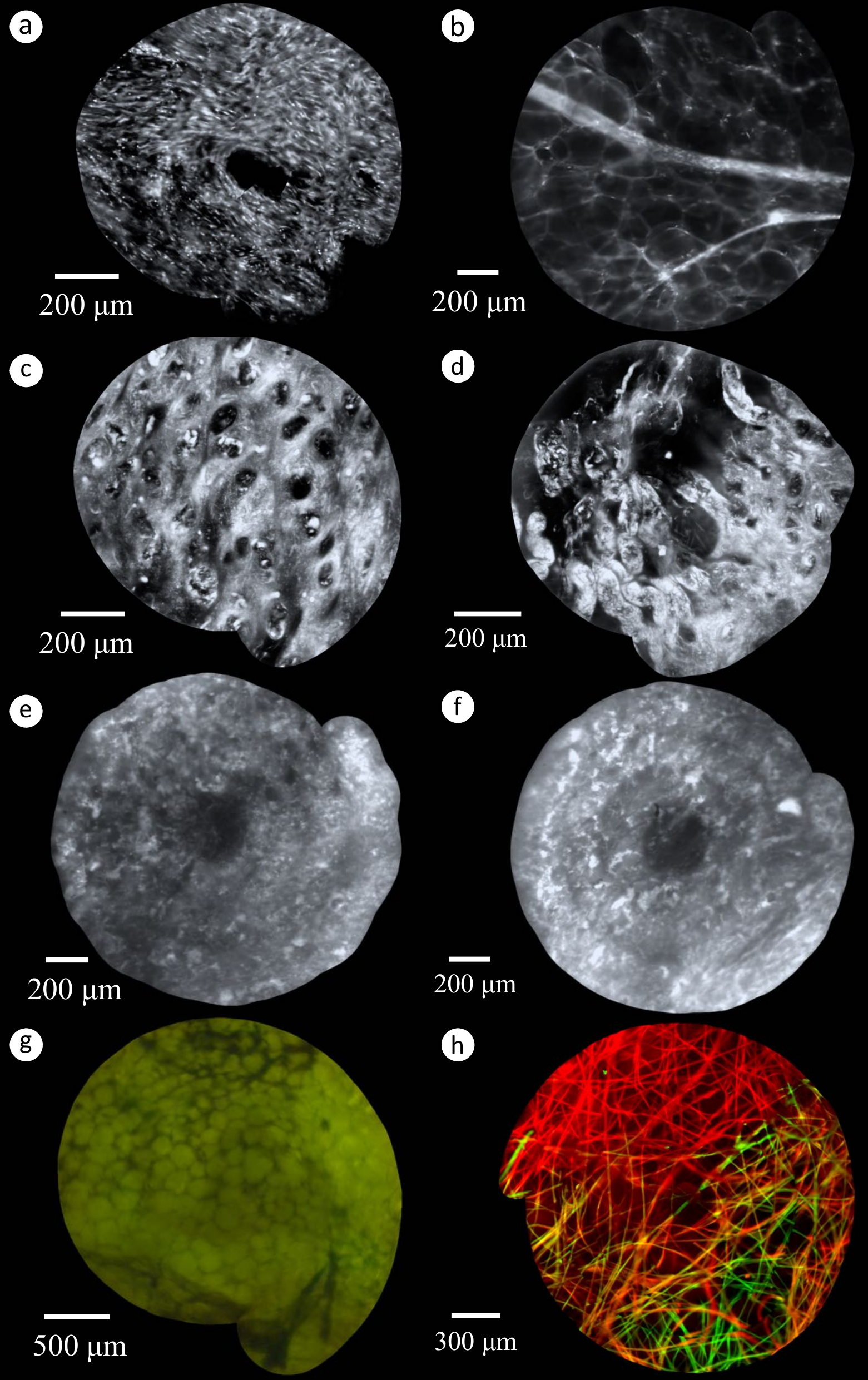}
\caption{Examples of \textit{ex vivo} tissue mosaics. (a) Human normal breast tissue (fibrous connective tissue), (b) human normal breast tissue (adipose tissue admixed with fibrous connective tissue), (c-d) porcine colonic tissue, and (e-f) human breast invasive carcinoma. (g-h) Example \textit{ex vivo} mosaic results using the in-house endocytoscopy with human normal breast tissue and the dual wavelength slit scanning confocal system with lens paper.\vspace{-0.5cm}}
\label{fig:Ex-vivo-tissue}
\end{figure}

\subsection{\textit{Ex Vivo} Mosaicking Results}

Mosaics generated from several samples of \textit{ex vivo} tissues and phantoms are shown in Fig.~\ref{fig:Ex-vivo-tissue} using different combinations of scanning speed, scan area and endomicroscopy system. Results in Fig.~\ref{fig:Ex-vivo-tissue}(a-f) were acquired using the single channel (488 nm) virtual slit endomicroscope, Fig.~\ref{fig:Ex-vivo-tissue}(g) was acquired using the fibre bundle endocytoscope, and Fig.~\ref{fig:Ex-vivo-tissue}(h) was acquired using the dual channel (488 nm and 660 nm) virtual slit endomicroscope. Fig.~\ref{fig:Ex-vivo-tissue}(a) and (b) are examples from \textit{ex vivo} human breast tissue stained with acriflavine, (c) and (d) are \textit{ex vivo} porcine colon tissue stained with acriflavine, (e) is \textit{ex vivo} human breast tissue exhibiting invasive carcinomas, (f) is an large area mosaic example using a lens tissue paper phantom stained with acriflavine, (g) is normal \textit{ex vivo} human breast tissue (adipose) stained with methylene blue, and (h) is a phantom of lens tissue paper stained at different points with a mix of methylene blue (generating the red pseudo-coloured fluorescence) and acriflavine (generating the green pseudo-coloured fluorescence). The diameters of the mosaics are approximately \SI{1}{\milli\metre} for Fig.~\ref{fig:Ex-vivo-tissue}(a,c-d), \SI{1.7}{\milli\metre} for Fig.~\ref{fig:Ex-vivo-tissue}(b,e), \SI{3.7}{\milli\metre} for Fig.~\ref{fig:Ex-vivo-tissue}(f), \SI{2.65}{\milli\metre} for Fig.~\ref{fig:Ex-vivo-tissue}(g) and \SI{2.2}{\milli\metre} for Fig.~\ref{fig:Ex-vivo-tissue}(h). Acquisition times were approximately \SIlist{7}{\second} for (a,b,e), \SIlist{13}{\second} for (c-d), \SIlist{77}{\second} for (f), \SIlist{46}{\second} for (g), and \SIlist{41}{\second} for (h), which varied due to different scan speeds, scan areas and frame rates.  

\subsection{Energy Delivery}

Finally, for the first time in robotic-assisted endomicroscopy, preliminary results showing how a fibre-delivered CO$_{2}$ laser ablation system can be integrated into the scanning device are presented. A device combining imaging and ablation could allow the surgeon to scan an area immediately prior to ablation, providing micro-scale image-guided intervention. The energy delivery could be utilised for direct therapy or for marking of tumour margins for subsequent resection by other means.

\begin{figure}[t]
\centering
\includegraphics[width=0.8\linewidth]{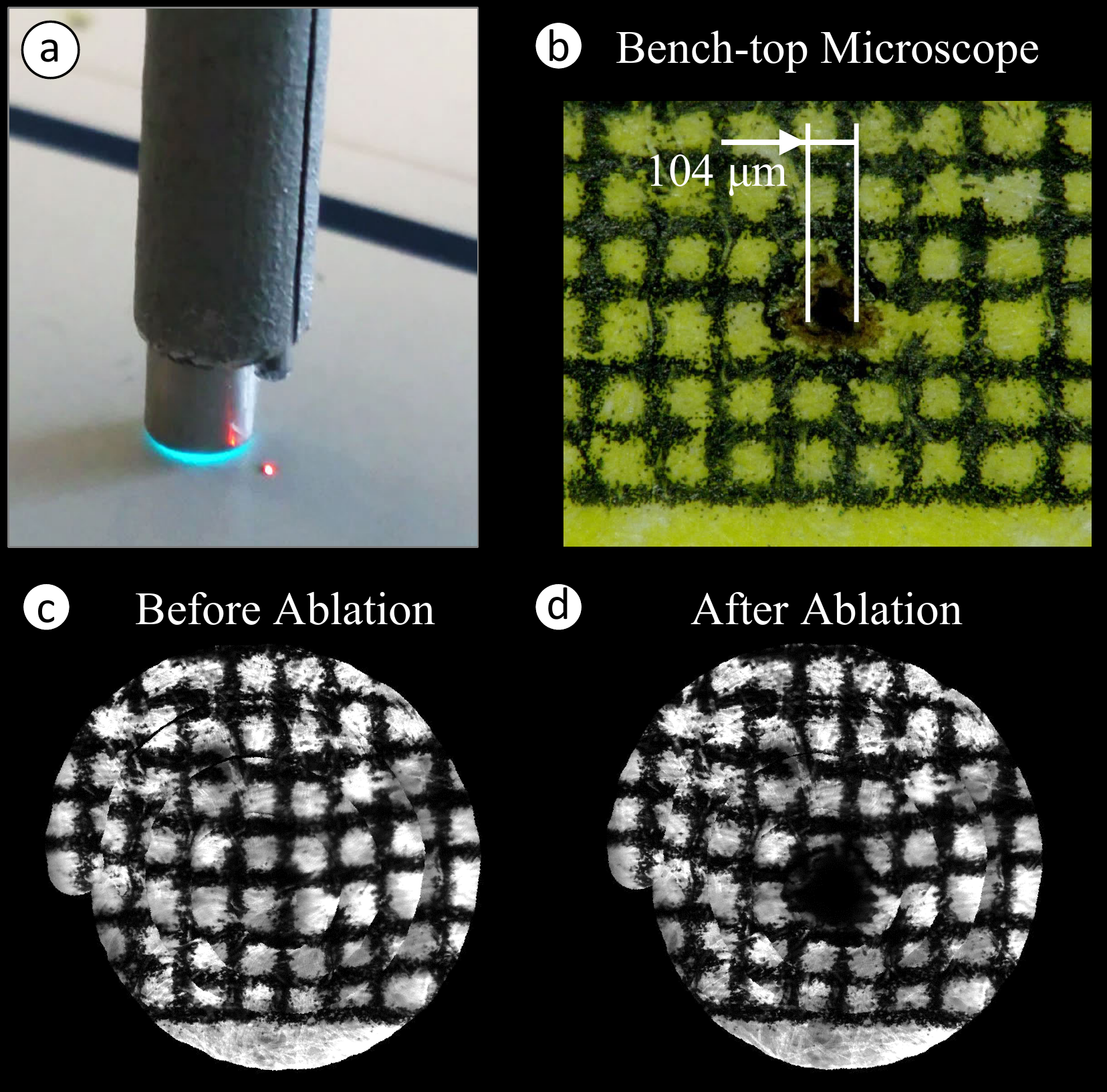}
\caption{(a) Instance from the laser ablation procedure on a paper card when the CO$_{2}$ laser fired. (b) A bench-top microscope image of the grid pattern, showing an ablated area and surrounding thermal spread with a radius of approximately \SI{210}{\micro\metre}. (c) and (d) show examples of a real-time mosaic image generated before and after the laser ablation, respectively. }
\label{fig:energyDelivery}
\end{figure}

The CO$_{2}$ laser (OmniGuide Inc., USA) was used in `Super Pulse' mode (single pulse, power \SI{3}{\watt} and duration \SI{40}{\milli\second}) in order to deliver a small precise mark onto the target. The laser fibre is inserted through the same channel as the microscopy fibre bundle and separated by a small horizontal mechanical offset at the tip of the instrument. The laser fibre was fixed with a vertical offset of \SIrange[range-units = single,range-phrase = --]{1}{2}{\milli\metre} from the tip of the microscopy probe to place it at the correct working distance when the endomicroscopy probe was in contact with the sample (see Fig.~\ref{fig:energyDelivery}(a)). Since the two fibres are offset laterally, in order to be able to ablate at the centre of the microscopy mosaic, a complimentary offset was applied to the probe's position during CO$_{2}$ laser ablation (see Supplementary Fig.~S8).

To demonstrate the instrument's ability to deliver laser energy at the centre of a mosaicked area, a spiral scan and mosaic was performed, the laser was fired, and the same area was re-scanned, as shown in Fig.~\ref{fig:energyDelivery}(b-c)  for comparison purposes. The laser delivered a very short pulse, ablating an area of \SI{104}{\micro\metre} diameter with minimal thermal spread (\SI{<50}{\micro\metre} radially), as can be seen in Fig.~\ref{fig:energyDelivery}(d). This result demonstrates the potential for the combination of microscopy with a laser ablation fibre in the same reference frame to assist with intraoperative tumour marking and ablation. There are mosaicking errors due to the lack of image signal in the ablated region, and this would become worse if this region was larger or off-centre. However, in practice, closed loop mosaicking would not be used for post-ablation inspection, although the user could employ either individual images or open loop mosaicking if needed. It should also be noted that ablation was only tested while the probe was not-scanning, and we do not envisage using the scanning mechanism to programme an ablation trajectory; the scanning mechanism is purely for imaging and for switching the position of the ablation fibre to the centre of the imaging field of view. 

\section{Discussion}

A novel robotic high-speed scanning device for intraoperative tissue identification and margin assessment has been demonstrated. The approach  adopted here was different to that of previous works in that it specifically targets clinical applications where a direct line of sight to the tissue of interest is available, such as in skin cancer, breast surgery or neurosurgery. Since for these applications the device is not required to bend or be flexible, it can generate fast, large area mosaics through a highly accurate and repeatable mechanism, while also making use of higher resolution endomicroscopy probes with large minimum radii of curvature. Combined with a high frame rate endomicroscope, the system can generate high resolution images over an area of \SI{3}{\milli\metre\squared} in approximately \SI{10}{\second}, whereas in previous reports the time to scan over a similar sized area was close to one minute.

The disadvantage of this new approach is that, while the rigidity of the instrument allows high resolution fibres to be inserted, it prevents its use at operating sites that require flexible access. Indeed, the access must allow the endomicroscopy probes to be placed approximately normal to the surface to obtain high quality microscopic images. Hand-held operation of the instrument is also challenging as the user is generally unable to maintain consistent contact forces between the probe and the tissue, and this force is not regulated or compensated by the device. The problem can be addressed by using the supporting robotic arm's hands-on approach to place and stabilise the instrument, but this adds significantly to the cost and complexity of the system. Hand-held or mounted use could be improved by incorporating a mechanism to axially control the tip of the instrument and maintain a constant contact force. This stabilisation mechanism could be active \cite{Latt2011a} or a simpler rigid stabilisation tube surrounding the scanning probe \cite{Erden2014b}.

Simple pairwise registration between images was used to allow real-time input to the visual servoing, with inevitable accumulation of errors. This approach will also not handle very large or abrupt motions, although the high frame rate endomicroscopy system used here is more tolerant to motion than slower systems. It may be possible to improve the performance of the mosaicking by making use of the programmed trajectory to reject spurious registrations, by estimating local non-rigid deformations, or by using external devices that track both the probe and the tissue; these are topics for further study.

Despite the need for a stabilising arm, the approach presents significant potential advantages over frozen section and other non-scanned optical biopsy techniques, as it significantly reduces the tissue assessment time. It is also minimally invasive,  requiring neither tissue removal nor destruction. Combined with interventional capabilities, shown for the first time in this work, this tool could not only provide real-time histological information but also mark and ablate suspicious areas and tumour margins intraoperatively, without the need for a separate tool. Finally, the small footprint of the system in the operating theatre, either as a hand-held instrument or combined with the small robotic arm, would allow it to be easily combined with other imaging modalities such as indocyanine green (ICG) fluorescence and hyper-spectral imaging, therefore creating a unified platform for intraoperative image guidance and therapy which includes real-time histological information.

\section{Conflict of Interest} 
\noindent There are no conflicts of interest.

\section{Acknowledgements}
\noindent Funding was provided by EPSRC Grants EP/I027769/1 and EP/N022521/1, SMART Endomicroscopy.

\ifCLASSOPTIONcaptionsoff
  \newpage
\fi

\end{document}